 \documentclass[final,5p,times,twocolumn]{elsarticle}

\usepackage{amssymb}
\usepackage{lipsum}

\usepackage[hidelinks]{hyperref}

\usepackage{amsmath}
\usepackage{titlesec}
\titleformat{\paragraph}[runin]
{\normalfont\normalsize\it}{\theparagraph}{1em}{}

\usepackage{soul}
\usepackage{xcolor}
\sethlcolor{red}

\usepackage{placeins}

\newcommand{\citeA}[1]{\citeauthor{#1} \cite{#1}}

\begin{document}

\begin{frontmatter}



\title{A Novel Unsupervised Machine Learning Strategy to Handle Multimodal Cardiac PET/MRI Data}

\author[affiliation_chu,affiliation_CRCI2NA]{Brunnhilde Ponsi}
\author[affiliation_chu,affiliation_CRCI2NA]{Thomas Carlier}
\author[affiliation_chu,affiliation_Inst_thorax_cardio]{Lara Marteau}
\author[affiliation_siemens]{Aurélien Monnet}
\author[affiliation_chu,affiliation_CRCI2NA]{Thomas Eugène}
\author[affiliation_chu,affiliation_Inst_thorax_radio]{Jean-Michel Serfaty}
\author[affiliation_chu,affiliation_Inst_thorax_cardio]{Nicolas Piriou}
\author[affiliation_chu,affiliation_CRCI2NA]{Hatem Necib\corref{cor1}}\ead{hatem.necib@chu-nantes.fr}

\affiliation[affiliation_chu]{organization={Nantes Université, CHU Nantes},
            addressline={F-44000 Nantes},
            country={France}}

\affiliation[affiliation_CRCI2NA]{organization={CRCI2NA},
            addressline={INSERM UMR 1307},
            city={Nantes},
            country={France}}

\affiliation[affiliation_Inst_thorax_cardio]{organization={Nantes Université, CHU Nantes, INSERM, Cardiology Department},
            addressline={INSERM UMR 1307,  CIC 1413, l’institut du Thorax},
            city={F-44000 Nantes},
            country={France}}
\affiliation[affiliation_siemens]{organization={Siemens Healthineers France},
            city={Courbevoie},
            country={France}}
            
\affiliation[affiliation_Inst_thorax_radio]{organization={Nantes Université, CHU Nantes, INSERM, Radiology Department},
            addressline={l’institut du Thorax},
            city={F-44000 Nantes},
            country={France}}

\cortext[cor1]{Corresponding author}


\begin{abstract}
Arrhythmogenic left ventricular cardiomyopathy is a genetic myocardial disease difficult to diagnose due to the lack of gold standard criteria. Simultaneous PET/MR imaging, combined with multiparametric quantitative analysis, could facilitate the identification of different profiles related to the phenotype and progression of cardiomyopathy. This preliminary study focuses on a methodological strategy for dealing with PET/MRI data, including inter-patient data linkage and regional analysis. 
Two-step clustering was applied to T1 and T2 maps, LGE, and 18F-FDG-PET images of 99 patients genetically diagnosed with arrhythmogenic left ventricular cardiomyopathy. Each patient’s images were independently z-scored and summed into a single volume, which was clustered into supervoxels. Thirty-two inter-patient groups of supervoxels were obtained by spectral clustering. An "abnormality" score was assigned to each cluster and modality, and used to visualise abnormal regions likely associated with disease. They enabled the generation of automated textual and bullseye health reports for each patient, which were compared with cardiac imager assessments using balanced accuracy in repeated nested cross-validation. This approach was further validated on a larger cohort of 167 numerical phantoms.
The reports generated by clustering accurately identified most of the cardiac physicians’ observations (BA = 0.76 $\pm$ 0.04 in repeated nested cross-validation on patients, and BA $\ge$ 0.8 on phantoms). Furthermore, the identified abnormal clusters closely matched their visual observations, facilitating the identification of varying degrees of fibrosis or inflammation on the images.
This approach enables a more systematic handling of multimodal PET/MRI data to characterise myocardial heterogeneity in arrhythmogenic left ventricular cardiomyopathy patients.
\end{abstract}


\begin{keyword}
 PET/MRI \sep Arrhythmogenic Cardiomyopathy \sep Clustering \sep Multimodal data \sep Patient profiles 


\end{keyword}

\end{frontmatter}




\section{Introduction}
\label{introduction}

Arrhythmogenic cardiomyopathy (ACM) is an inherited myocardial disease characterized by fibrofatty replacement of the ventricular myocardium. Since 2010, it has emerged that ACM, which was originally believed to affect only the right ventricle (RV) can also affect the left ventricle (LV), either completely or partially with isolated LV fibrosis and normal LV function. In 2020, the Padua criteria were introduced to redefine ACM and its diagnostic criteria \cite{Corrado_Padoue_criteria_2020, th_Corrado_ACM_diagnostic_criteria_2024}. These criteria propose a multi-parametric approach to diagnosis, including specific criteria for arrhythmogenic left ventricular cardiomyopathy (ALVC). They emphasise the importance of cardiac MRI (CMR), which is now considered a necessary diagnostic tool as it allows tissue characterization by late-gadolinium enhancement (LGE) as a major ACM diagnostic criterion \citep{th_Corrado_ACM_diagnostic_criteria_2024}.

LGE techniques are extremely valuable in assessing myocardial fibrosis, but they are limited by the fact that they are qualitative, which reduces their usefulness as an inter-patient diagnostic tool. Additionally, they fail to address diffuse fibrosis in the cardiac muscle, which is an early indicator of disease that could be mitigated if detected early. Therefore, using alternative sequences such as T1 and T2 mapping is beneficial, as it provides additional quantitative information on the presence of diffuse fibrosis or edema \citep{Everett_assessment_myocardial_fibrosis_with_T1_2016}.

Although CMR is the reference diagnostic tool for identifying myocardial lesions, it is often inadequate for detecting inflammation outside the acute phase of cardiomyopathy \citep{th_tessier_PET_for_detection_of_myocardial_inflammation_in_ALVC}.  However, some patients with ACM may exhibit myocardial inflammation, which is a poor prognostic marker for cardiomyopathy, although treatment can improve event-free survival. Studies have shown the value of FDG-PET in identifying markers of myocardial inflammation, which is prevalent in a significant proportion of patients with ALVC \citep{th_tessier_PET_for_detection_of_myocardial_inflammation_in_ALVC, th_neves_cardiac_PET_in_ACM}. These studies demonstrated the complementary role of FDG-PET compared with cardiac MRI, highlighting the need for further investigations into the impact of using PET to characterize and diagnose ACM \citep{th_Protonotarios_role_FDG_PET_in_ACM}.

Then, while CMR provides valuable anatomical insights and information on fibrosis and edema, FDG-PET can bring complementary functional information. Acquiring PET and MRI simultaneously using a PET-MRI system thus has the potential to significantly improve ACM diagnosis. Multi-parametric PET-MRI may deepen our understanding of ACM, providing a more precise and straightforward diagnosis. It could also lead to the identification of different patient profiles with advanced cardiomyopathy, enabling the development of targeted therapeutic strategies that will be unique to each patient.

In particular, double-step clustering approaches based on unsupervised machine learning techniques applied in two successive stages have already been successfully utilized in several oncology studies for prediction, prognosis, and tumor segmentation \citep{th_inano_visualization_heterogeneity_regional_grading_gliomas_by_multiple_features_using_MRI_clustered_images, th_tatekawa_differentiating_IDH_status_gliomas_ML_MR-PET, th_even_clustering_multiparametric_functional_imaging_to_identify_high_risk_subvolumes_lung_cancer, th_hansen_unsupervised_supervoxel-based_lung_tumor_segmentation_PET-MR}. Supervoxel clustering offers a balance between data analysis methods disregarding spatial context by averaging and pixel-wise analyses, which are both computationally costly and vulnerable to noise and registration inaccuracies \citep{th_even_clustering_multiparametric_functional_imaging_to_identify_high_risk_subvolumes_lung_cancer, th_hansen_unsupervised_supervoxel-based_lung_tumor_segmentation_PET-MR}.

In this study, we propose a two-step clustering approach to handle multimodal PET-MRI volumes from ALVC patients.  This method automatically identifies all possible combinations of high and low states across the multiple modalities and detects the corresponding regions of interest. The approach involves grouping voxels from each patient into supervoxels, followed by inter-patient clustering of the resulting supervoxels. A key innovation of our method is the introduction of a scoring system that not only characterizes cluster-level abnormalities but also bridges the gap to clinical practice through automated report generation. These clusters have the potential to reveal new ALVC biomarkers that could help determine new phenotypes and identify different patient profiles. The method was quantitatively evaluated on patient data using a nested cross-validation framework. It was further validated on phantoms and illustrated through a representative patient example.

\section{Method}
\subsection{Database}
\begin{figure}[!t]
\centering
\includegraphics[width=1\linewidth]{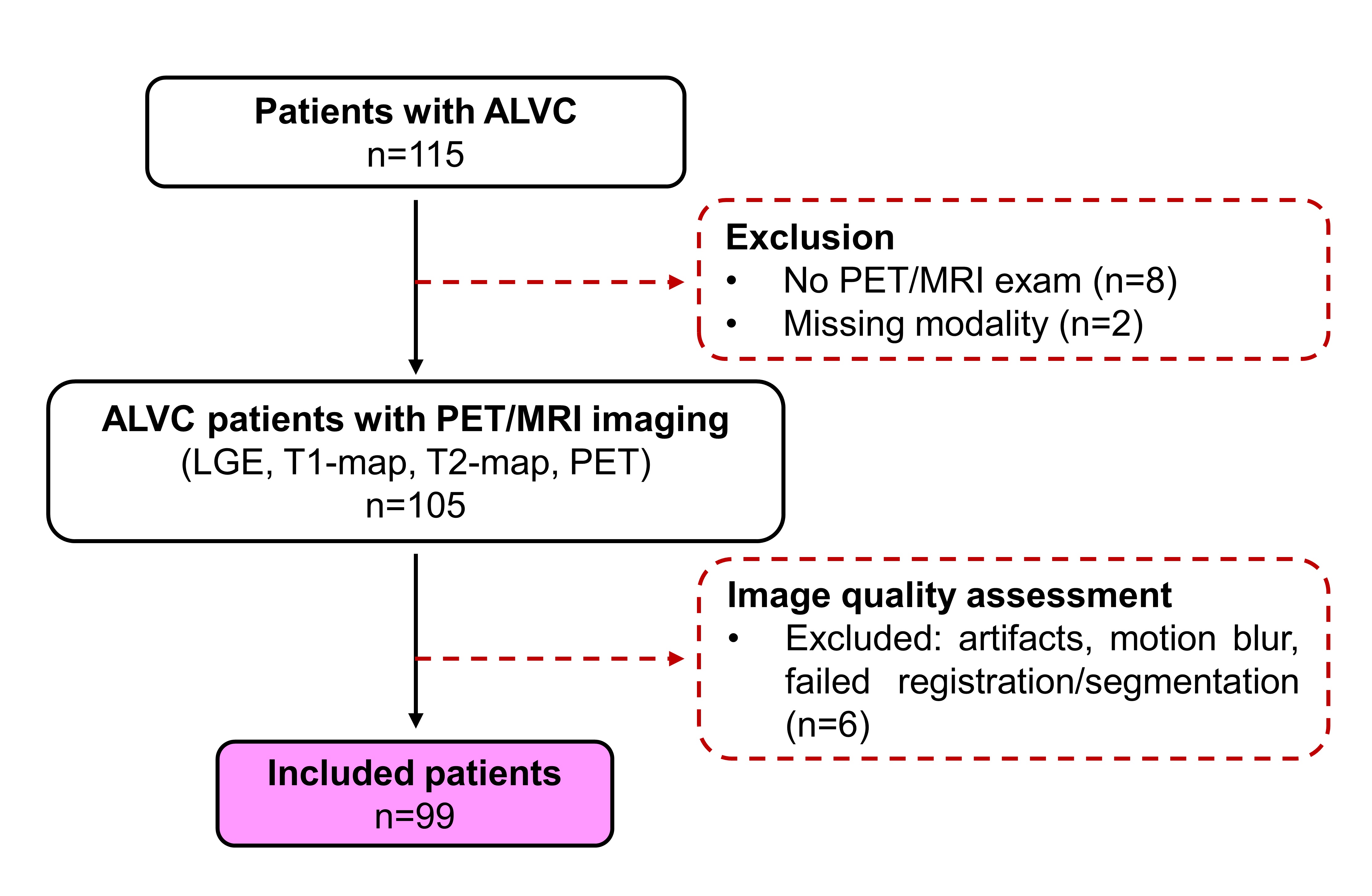}
\caption{Study flowchart.}
\label{fig_1_flowchart}
\end{figure}

The database comprised data from 115 patients who were diagnosed with an ACM genetic variant. All patients were enrolled in the prospective trial "CharACTPET-MR" (NCT05450783). PET-MRI data were  acquired using the Siemens PET-MR mMR Biograph. Analysis was performed using quantitative T1 and T2 maps, PET images with attenuation correction (PET-AC), and LGE-PSIR short-axis acquisitions. As described in Fig. \ref{fig_1_flowchart}, the final cohort comprised 99 patients following the exclusion of those with missing or poor-quality data. 

\subsection{Workflow of the method}
\begin{figure}[!t]
\centering
\includegraphics[width=1\linewidth]{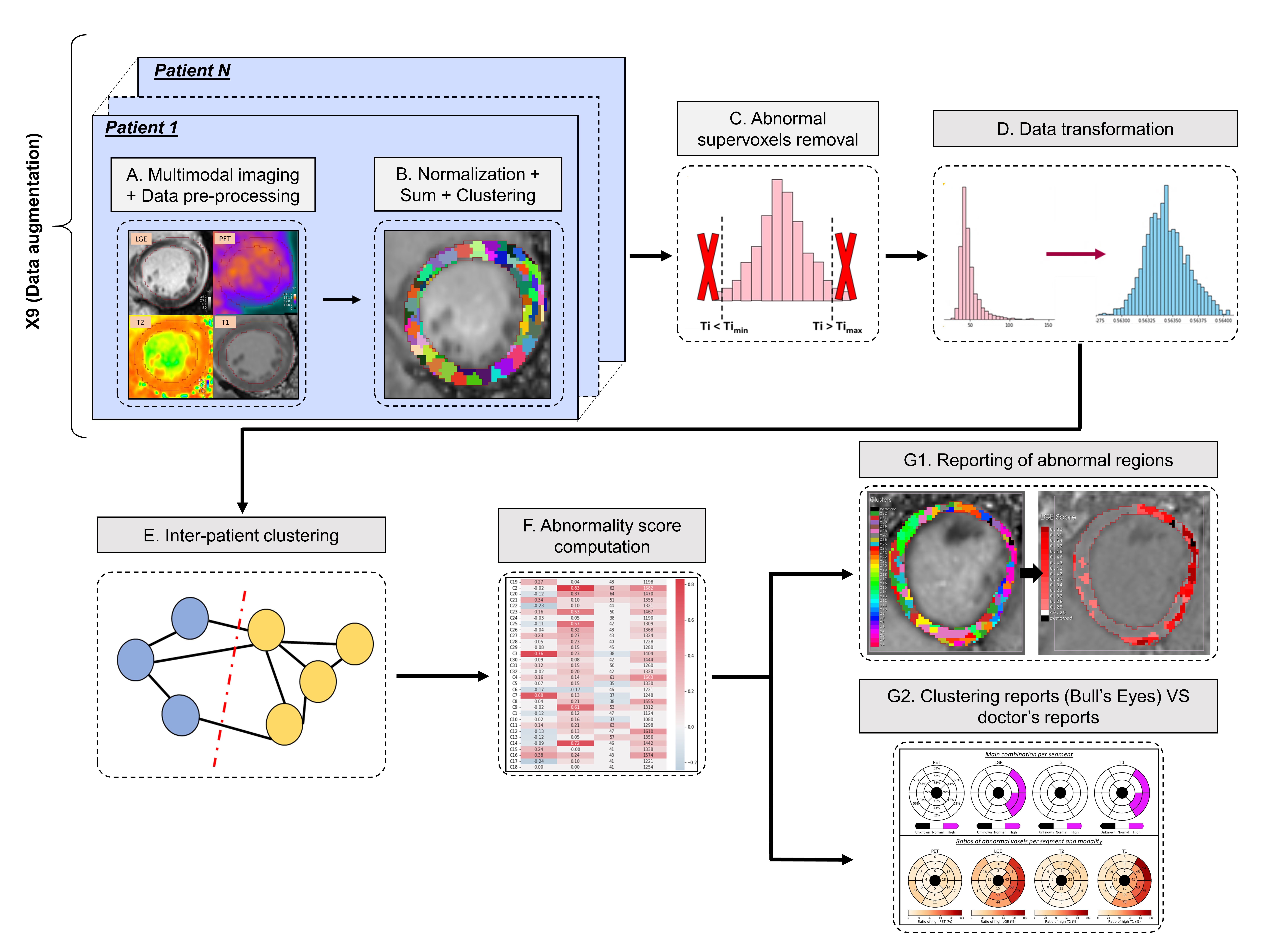}
\caption{Workflow of the two-step multimodal PET-MR imaging clustering.}
\label{fig_2_workflow}
\end{figure}

The workflow of this two-step clustering method is summarized in Fig. \ref{fig_2_workflow} by eight steps from \textit{A} to \textit{G2}.

\subsubsection{Step A. Data pre-processing and preparation}\label{met_stepA}
\begin{figure}[!t]
\centering
\includegraphics[width=1\linewidth]{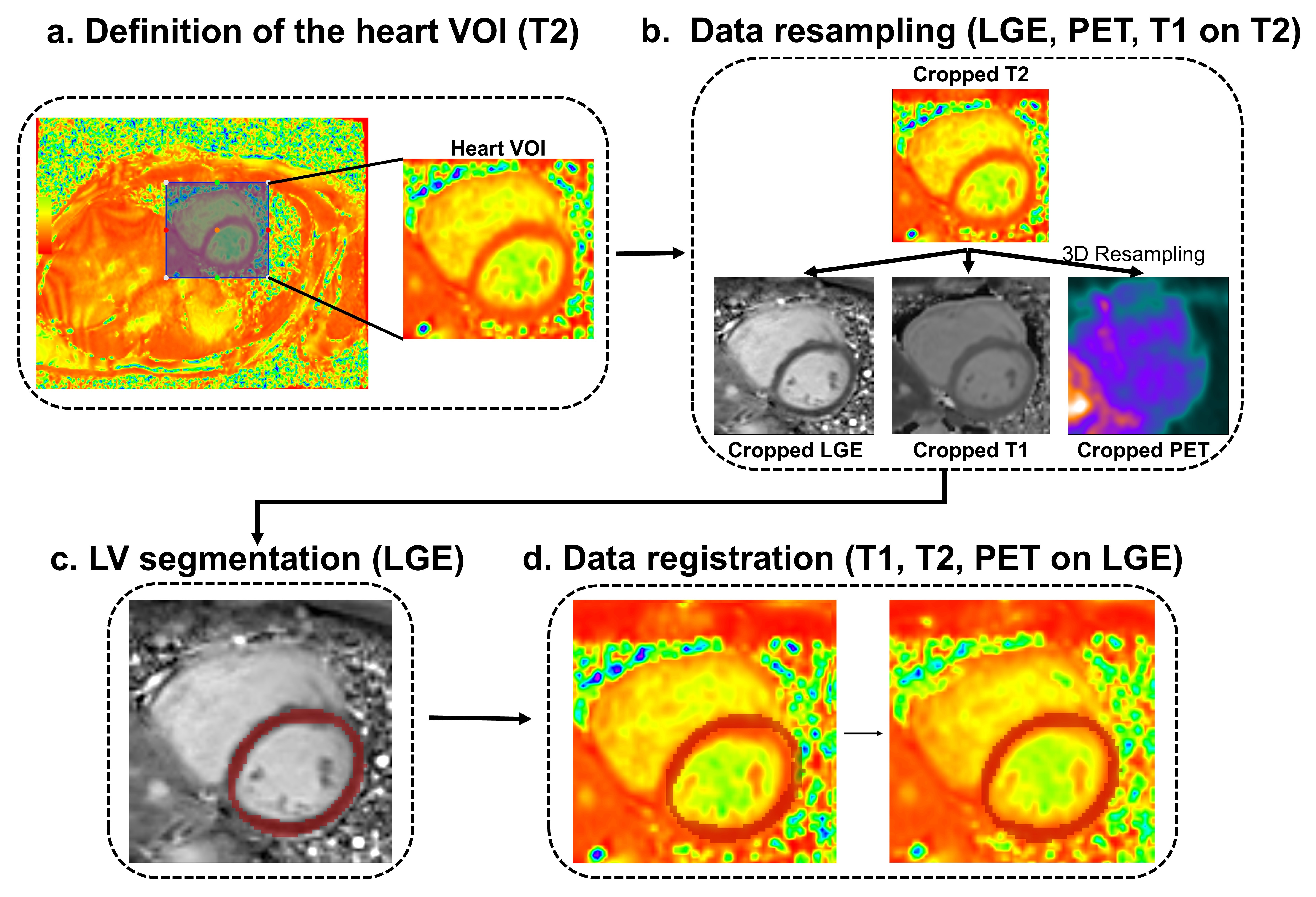}
\caption{Workflow of the data pre-processing steps (Fig. \ref{fig_2_workflow} Step A).}
\label{fig_3_preprocessing_steps}
\end{figure}

Data was pre-processed (Fig. \ref{fig_2_workflow} Step A) for each patient individually using the\textit{ 3D Slicer} software \cite{th_3D_slicer, th_3D_slicer_computing_platform}, as described in the next paragraphs and summarized in  Fig. \ref{fig_3_preprocessing_steps}.

\paragraph{Definition of the region of interest (ROI): }
 
For each patient, a volume of interest (VOI) was manually defined on the T2 or T1 map. This VOI corresponds to a rectangular volume surrounding the entire heart, as in  Fig. \ref{fig_3_preprocessing_steps} Step a, and was used for the resampling and registration steps.

\paragraph{Data resampling: }
 
All other sequences were resampled in all 3 dimensions on the heart VOI with BSpline interpolation (Fig. \ref{fig_3_preprocessing_steps} Step b).

\paragraph{Segmentation of the left ventricle (LV):}
 
The left ventricle was semi-automatically segmented on the LGE images (Fig. \ref{fig_3_preprocessing_steps} Step c) by a cardiologist on the Circle software by CVI \cite{th_circle_software_cvi}. Segmentations of the epicardium and endocardium were then exported and merged into a single LV segmentation.

\paragraph{Data registration:}
 
Although all MRI sequences were acquired at the same time point in the cardiac cycle, misalignment between MRI images can occur due to cardiac, respiratory, and patient movement. Additionally, since PET images are averaged over multiple cardiac cycles, they are not always perfectly aligned with MRI images. All modalities were then registered on the LGE images (Fig. \ref{fig_3_preprocessing_steps} Step d), either using the Elastix module \cite{Klein_elastix_2010} implemented in 3D Slicer or via manual registration if registration failed. All registrations were validated by a senior radiologist.

\subsubsection{Step B. First clustering: Intra-patient clustering, from voxels to supervoxels}\label{met_stepB}
A first clustering phase (Fig. \ref{fig_2_workflow} Step B) was used for each patient to group adjacent and similar voxels into supervoxels. This reduced registration issues, computational complexity, and aberrant voxel values.

This initial clustering was applied on a new ROI surrounding the LV, as in the example of Fig. \ref{fig_4_results_example_visualization}. The mean value of each of the four PET or MRI modalities within this ROI was subtracted, and the result was divided by the standard deviation to ensure equal contribution from all modalities. These z-scores were then summed to create a single cumulative “Z-volume”. Supervoxels were extracted from this Z-volume using SLIC (a k-means-based algorithm) \cite{Achanta_SLIC_2012}. The chosen parameters are described in \ref{app1:SLIC_param}. 

The LV segmentation mask was applied to select only supervoxels containing at least 20\% of their volume within the LV, thereby eliminating those caused by registration errors (over-segmentation). Each supervoxel was assigned T1, T2, LGE and PET values corresponding to their median pixel values within the LV. All data outside the LV segmentation was masked. While the T1 and T2 values remained quantitative, only the z-score values of each supervoxel were saved for the LGE and PET values. In addition, the position of each supervoxel within the 17 cardiac segments was recorded. Where a supervoxel contained voxels spanning several cardiac segments, the segment containing the majority of the supervoxel's voxels was selected. 

Finally, standard data augmentation techniques (rotations and flips) were used on steps A and B to reduce the sensitivity of the method to imaging variations as described in \ref{app2:data_augmentation}. 

\subsubsection{Step C. Removal of supervoxels with medically impossible relaxation times}
Both longitudinal (T1) and transversal (T2) relaxation times have been well-established in healthy myocardium \cite{th_von_Knobelsdorff_myocardial_T1_T2_map_3T_2013}. Some studies have also investigated elevated T1 and T2 values in diseased myocardium. Knowledge of these values makes it possible to establish thresholds above or below which relaxation times are deemed implausible in the myocardium. Such values could be attributed to factors such as fat or MRI artifacts, rendering them unsuitable for consideration in our study. Consequently, extreme (minimum and maximum) values were defined for T1 and T2, and any supervoxel with T1 or T2 values exceeding one of these thresholds was excluded from subsequent analyses (Fig. \ref{fig_2_workflow} Step C). Plausible T1 and T2 values for the myocardium were defined as follows:
\begin{subequations}\label{eq_plausible_T1_T2}
\begin{align}
    1000 \text{ ms} \leq \text{T1}_{\text{myocardium}} &\leq 1800 \text{ ms} \\
    20 \text{ ms} \leq \text{T2}_{\text{myocardium}} &\leq 80 \text{ ms}
\end{align}
\end{subequations}

\subsubsection{Step D. Data transformation}
As data in its original form is not always optimal for statistical analysis, it is common practice to re-express the data in a modified format that is more suitable for analysis. These modified representations can exhibit characteristics such as increased dispersion and improved symmetry, or simply become more interpretable and manipulable. Power transformations, in particular, are widely used in literature due to their simplicity and favorable properties, such as continuity, differentiability, and preservation of data rank \cite{th_stoto_power_transformation}. 

Of these transformations, the Box-Cox transformation \cite{th_boxcox} is widely used to normalize feature distributions by reducing skewness and stabilizing variance. By making the data more Gaussian-like, it improves the behavior of distance-based metrics and enhances the performance of clustering algorithms, ultimately leading to better-defined clusters \cite{th_hansen_unsupervised_supervoxel-based_lung_tumor_segmentation_PET-MR}. The Box–Cox transformation was therefore applied to the pooled supervoxels from all patients in an adapted version that included data with negative values (Fig. \ref{fig_2_workflow} Step D). The modified form is expressed as follows \cite{th_boxcox}:
\begin{equation} \label{eq_boxcox}
    y^{(\lambda)} =
    \begin{cases}
        \frac{(y+\lambda_2)^{\lambda_1}-1}{\lambda_1} & \text{if } \lambda_1 \ne 0\\
        \log(y+\lambda_2) & \text{if } \lambda_1 = 0\\
    \end{cases}
\end{equation}
where $y$ is the combined supervoxels data before transformation,  $y^{(\lambda)}$ is the transformed data, $\lambda_2$ is chosen so that $y_i+\lambda_2 > 0$ for all $y_i \in y $ ,  and $\lambda_1$ is defined to maximize the log-likelihood, to approach a normal distribution of the transformed data.
We defined $\lambda_1$ and $\lambda_2$ independently for each modality, on the combined supervoxels data of all the patients, and we took $\lambda_2^{(mod)} = |{\min{y_i^{(mod)}}}| + 10^{-5}$ to ensure positivity in the logarithm.

\subsubsection{Step E. Second clustering: Inter-patient clustering of the supervoxels}
Spectral clustering was used on Box-Cox transformed data to group all supervoxels into inter-patient clusters (Fig. \ref{fig_2_workflow} Step E). This method was chosen because it produced good results in our context and performed better than hierarchical clustering in the presence of noise, as has already been demonstrated \cite{th_hansen_unsupervised_supervoxel-based_lung_tumor_segmentation_PET-MR}. To facilitate the construction of the spectral clustering matrix, duplicate supervoxels were retained only once during the clustering process.

Various sets of parameters were tested to group the supervoxels into clusters, as described in the later sections.

\subsubsection{Step F. Abnormality score}
After the supervoxels were grouped into clusters, the statistical significance of these clusters was evaluated, and a score was attributed to each cluster and modality (Fig. \ref{fig_2_workflow} Step F). This was done to identify abnormal regions and compare them with physicians’ reports.

\paragraph{ANOVA and Tukey tests:} To assess significant differences between clusters, an ANOVA test was conducted to examine the substantial differences in cluster medians for each modality independently. If the difference was significant ($p < 0.05$), a Tukey test was performed for each modality to establish statistical differences between each pair of clusters. We then examined the clusters that were statistically different from the reference cluster, as described in the following sections.

\paragraph{Choice of a reference cluster:}
  
As LGE and PET values are not strictly quantitative, a reference cluster had to be defined. This reference cluster, $C_p$, comprised supervoxels that were presumed to be healthy, and was used as a baseline for comparison with the median values of the other clusters. As all subjects presented with ALVC, this cluster of healthy supervoxels had to be identified among the patients' supervoxels. 

The reference cluster was defined as the “healthiest” among “valid” clusters, which were defined as those that met the inclusion criteria indicative of healthy tissue. “Valid clusters” had to contain  more supervoxels than the average number of supervoxels per cluster 
and to exhibit median relaxation times within the normal ranges ($1050 \text{ ms} < T1 < 1300 \text{ ms}$; $38 \text{ ms} < T2 < 50 \text{ ms}$) \cite{th_von_Knobelsdorff_myocardial_T1_T2_map_3T_2013}. Additionally, the median LGE and PET values had to exceed the $5^{\text{th}}$ percentile of their respective global distributions, while the median PET value remained below the $60^{\text{th}}$ percentile. Within this subset of valid clusters, the reference cluster was defined as the one with the lowest mean LGE value, in accordance with the Padua criteria that identify LGE hypersignal as the main marker of ACM. These constraints ensured that $C_p$ could not correspond to a small region or be characterized by LGE/PET hyposignal or abnormally high PET activity.

\paragraph{Scoring of the clusters for each modality: }
  
Each cluster was assigned an abnormality score for each modality. The method of determination of this score varied based on the modality under consideration. 

To assess the PET or LGE abnormality of a cluster $C_i$, the score $\text{s}_{C_i}^{mod}$ was calculated relative to the median value $x_{C_p}^{mod}$ of the reference cluster $C_p$ using the formula: 
\begin{equation}\label{eq_score_TEP_LGE}
    \text{s}_{C_i}^{mod} = \frac{x_{C_i}^{mod} - x_{C_p}^{mod}}{x_{max}^{mod}-x_{min}^{mod}}
\end{equation}
where $mod$ is the considered modality between PET or LGE, $x_{C_i}^{mod}$ is the mean value of the considered cluster for the considered modality, and $x_{max}^{mod}$ and $x_{min}^{mod}$ respectively are the maximum and minimum mean values within the clusters in the considered modality. 

The T1 and T2 abnormality scores of a cluster were quantitatively assessed, based on the myocardial T1 and T2 values at 3T given by  Von Knobelsdorff-Brenkenhoff \cite{th_von_Knobelsdorff_myocardial_T1_T2_map_3T_2013}, and confirmed by senior cardiologists. The following definitions were used, based on a piecewise affine function defining a zero abnormality score for T1 and T2 values within clinically defined healthy ranges:
\begin{equation} \label{eq_score_T1}
    \text{s}_{C_i}^{T1} =
    \begin{cases}
        -0.99 & \text{if } T1 < 1000 \text{ ms} \\
        \frac{T1_{C_i} - 1050 - 100}{1000} & \text{if } 1000 \text{ ms} \leq T1 < 1050 \text{ ms} \\
        0 & \text{if } 1050 \text{ ms} < T1 < 1300 \text{ ms} \\
        \frac{T1_{C_i} - 1300 + 100}{1000} & \text{if } 1300 \text{ ms} \leq T1 < 1800 \text{ ms} \\
        1 & \text{if } T1 > 1800 \text{ ms}
    \end{cases}
\end{equation}

\begin{equation}\label{eq_score_T2}
    \text{s}_{C_i}^{T2} =
    \begin{cases}
        -0.99 & \text{if } T2 < 20 \text{ ms} \\
        \frac{T2_{C_i} - 38 - 10}{100} & \text{if } 20 \text{ ms} \leq T2 < 38 \text{ ms} \\
        0 & \text{if } 38 \text{ ms} < T2 < 50 \text{ ms} \\
        \frac{T2_{C_i} - 50 + 10}{100} & \text{if } 50 \text{ ms} \leq T2 < 100 \text{ ms}  \\
        1 & \text{if } T2 > 100 \text{ ms}
    \end{cases}
\end{equation}
where $T1_{C_i}$ and $T2_{C_i}$ are the mean T1 and T2 values of cluster $C_i$.

Clusters not statistically different from the reference cluster in a given modality, based on the Tukey test, were assigned a score of 0 for that modality.

\subsubsection{Step G1. Visualisation of the abnormal clusters on the images: }
  
To visualise the pathological zones detected on the medical images (Fig. \ref{fig_2_workflow} Step G1), the computed abnormality scores were superimposed on the original MRI and PET images. This highlighted areas with elevated LGE, T1, T2, or PET values. Supervoxels that had been previously excluded due to implausible T1 or T2 values were also superimposed on the original images to aid the identification of regions affected by poor registration, artifacts, or fat. 

\subsubsection{Step G2. Quantitative report generation and evaluation}

\paragraph{Obtaining a medical report for each patient:}
  
In order to summarise the cluster-derived information succinctly in a clinically useful format and to assess the performance of the clustering in relation to the diagnoses of cardiologists and radiologists, the cluster scores were converted into reports that are comparable with those produced by cardiac imagers (Fig. \ref{fig_2_workflow} Step G2). 

Each cluster was categorized as having normal or high values for each of its four modalities, based on thresholding of the cluster's score within each modality. The threshold value was set at 0.1 for both T1 and T2 since the formulation of their scores was designed to define 0.1 as the upper boundary of normality. Since the PET and LGE scores were defined as distances to the reference cluster, their score thresholds were defined as hyperparameters in the nested cross-validation described in the next section.

Subsequently, each cluster was associated with one of two categories for each of its four imaging modalities: high or normal. Patients were then assigned to the proportion of their cluster combinations, resulting in a personalized report detailing the distribution of various regions within their myocardium.

\paragraph{Evaluation of the reports compared to the physicians’ reports:}
   
The method was evaluated by comparing the PET/LGE/T1/T2 combinations identified by clustering with those observed by physicians. 

To ensure consistency with the clustering-based report, physicians assessed each myocardial segment, noting the presence of high-intensity voxels in each of the four imaging modalities.

A repeated nested cross-validation framework was used to compare the results. The score was defined as the mean balanced accuracy (BA) across all patients, averaged across all outer folds and repetitions. The dataset was split into three outer folds, each of which was subjected to a grid search with two-fold inner cross-validation. The grid search process included multiple sets of hyperparameters, including variations in the spectral clustering parameters and the high-threshold values for PET and LGE.

\paragraph{Representation of the clustering reports with Bull’s eye plots: }
The clustering reports were summarized using two types of Bull’s Eye plot. The first type represented the main PET/LGE/T1/T2 combination observed per segment. The second type showed the percentage of high modality per segment.

First, the Bull's eye plots of the main combination per segment show the proportion of main imaging combinations (formed by normal or abnormally high T1, T2, LGE, and PET values) for each patient and segment. If a modality has a high value in a combination, it is represented in pink; otherwise, it is represented in white. This approach enables specific myocardial tissue to be characterized based on multimodal imaging.

Secondly, the Bull's eye plots of the ratios of abnormally high value voxels per segment per modality illustrate the ratio of high-value voxels per segment and modality, offering a quantitative perspective on the distribution of abnormalities within cardiac segments.

\subsection{Validation of the method on numerical phantoms}
The clustering method was finally validated using a set of 167 numerical phantoms that simulated LGE, PET, T1, and T2 cardiac volumes. The process of creating the phantom is described in \ref{app3:phantoms}. Double-step clustering was applied to the phantoms using the same parameters selected for the patients through nested cross-validation. The only difference was that the number of SLIC supervoxels was limited to 1000, due to the simpler anatomical representation of the phantoms and computational considerations. Various levels of noise $\eta$ were tested: $\eta\in{0\%, 5\%, 10\%}$.

\FloatBarrier
\section{Results}
\subsection{Quantitative evaluation of the method}
The model applied to the ninety-eight patients achieved a mean balanced accuracy of $0.76 \pm 0.04$ in repeated nested cross-validation across three outer folds and 56 repetitions.
	
To further validate the method, it was applied to the 167 phantoms, using the hyperparameters chosen via nested CV (\ref{app4:hyperparameters}). Balanced accuracies of 0.88, 0.80, and 0.80 were obtained, respectively, when the phantoms were created with noise levels of 0, 5\%, and 10\% (\ref{app3:phantoms}).

\subsection{Qualitative evaluation of the method}
 
To qualitatively evaluate the abnormal PET and MRI regions detection, the model was finally retrained on the full patient dataset using the chosen hyperparameters, giving a balanced accuracy of 0.81, with a mean sensitivity and specificity of 0.76 and 0.86. 

\begin{figure}[!t]
\centering
\includegraphics[width=1\linewidth]{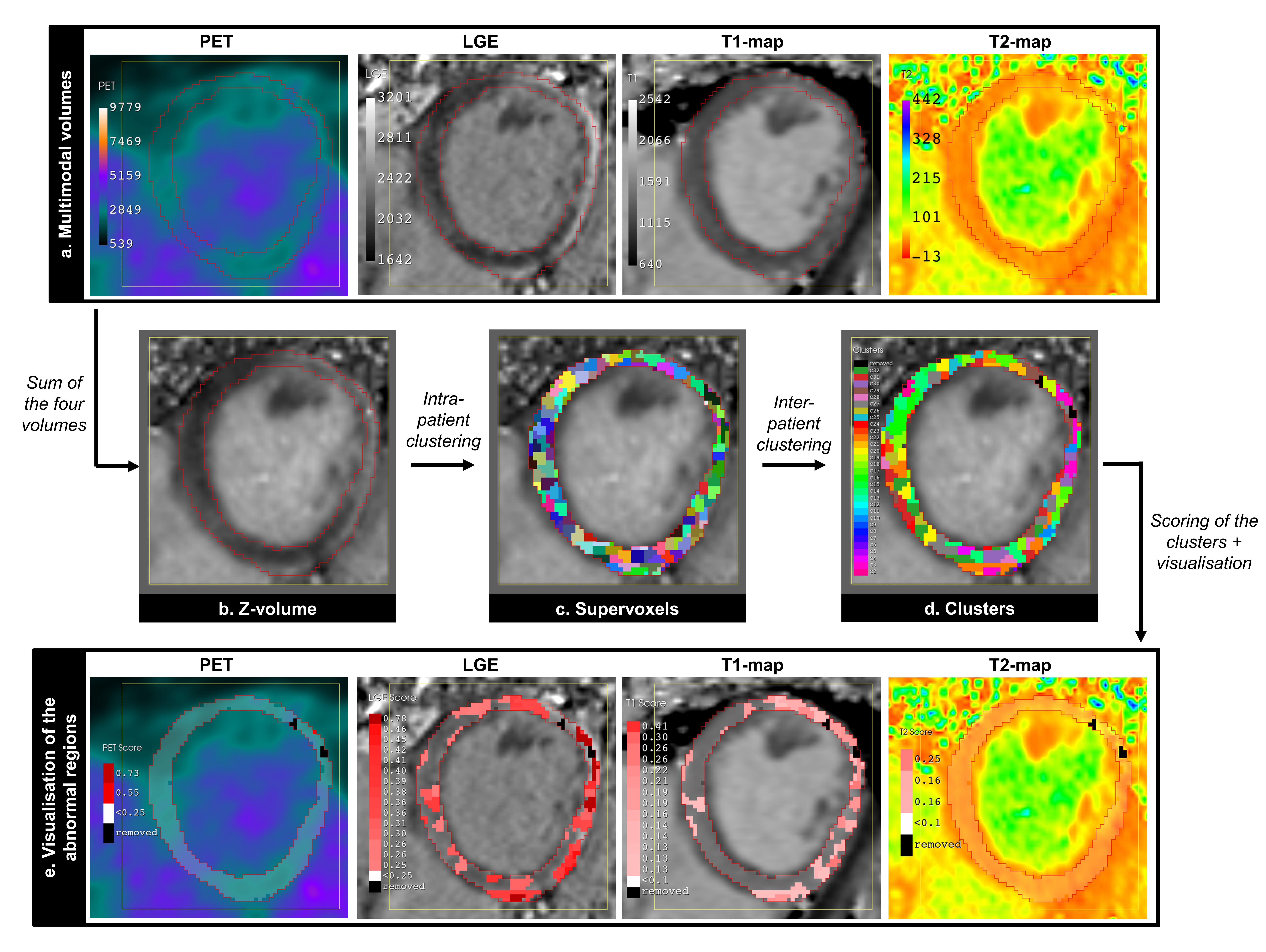}
\caption{Automatic detection workflow of the abnormal PET and MRI regions, illustrations on one basal slice of a patient from the cohort. The yellow square defines the LV VOI on which the clustering steps were applied. The multimodal volumes (a) are normalized and summed to one single ‘Z-volume’ (b), on which intra-patient clustering is applied to identify supervoxels (c). Supervoxels from all patients are then combined to identify 32 clusters that can be reported to the images (d). Lastly, the clusters are scored based on the intensity of their voxels, and clusters with high scores can be reported on the images to visualise abnormal regions for each modality (e).
}
\label{fig_4_results_example_visualization}
\end{figure}

This two-step clustering workflow automatically highlights regions of interest in multimodal images by detecting areas associated with abnormal scores. By performing inter-patient clustering on normalized supervoxels, the method may reveal subtle pathological patterns invisible to visual inspection, which relies primarily on contrast differences. Fig. \ref{fig_4_results_example_visualization} presents an example of this automated detection workflow using a basal LGE slice from a representative patient.

As shown in Fig. \ref{fig_4_results_example_visualization}, the patient exhibits complete extinction of physiological myocardial fixation and no hypermetabolism in the left ventricle. The visualisation method highlighted no high voxel values for both PET and T2-map. However, it detected a large area of late contrast uptake under the epicardium of the lateral wall. This same area exhibits T1 values greater than 1300 ms and was highlighted by clustering.

\subsection{Automatisation of the multimodal image analysis}
\begin{figure}[!t]
\centering
\includegraphics[width=1\linewidth]{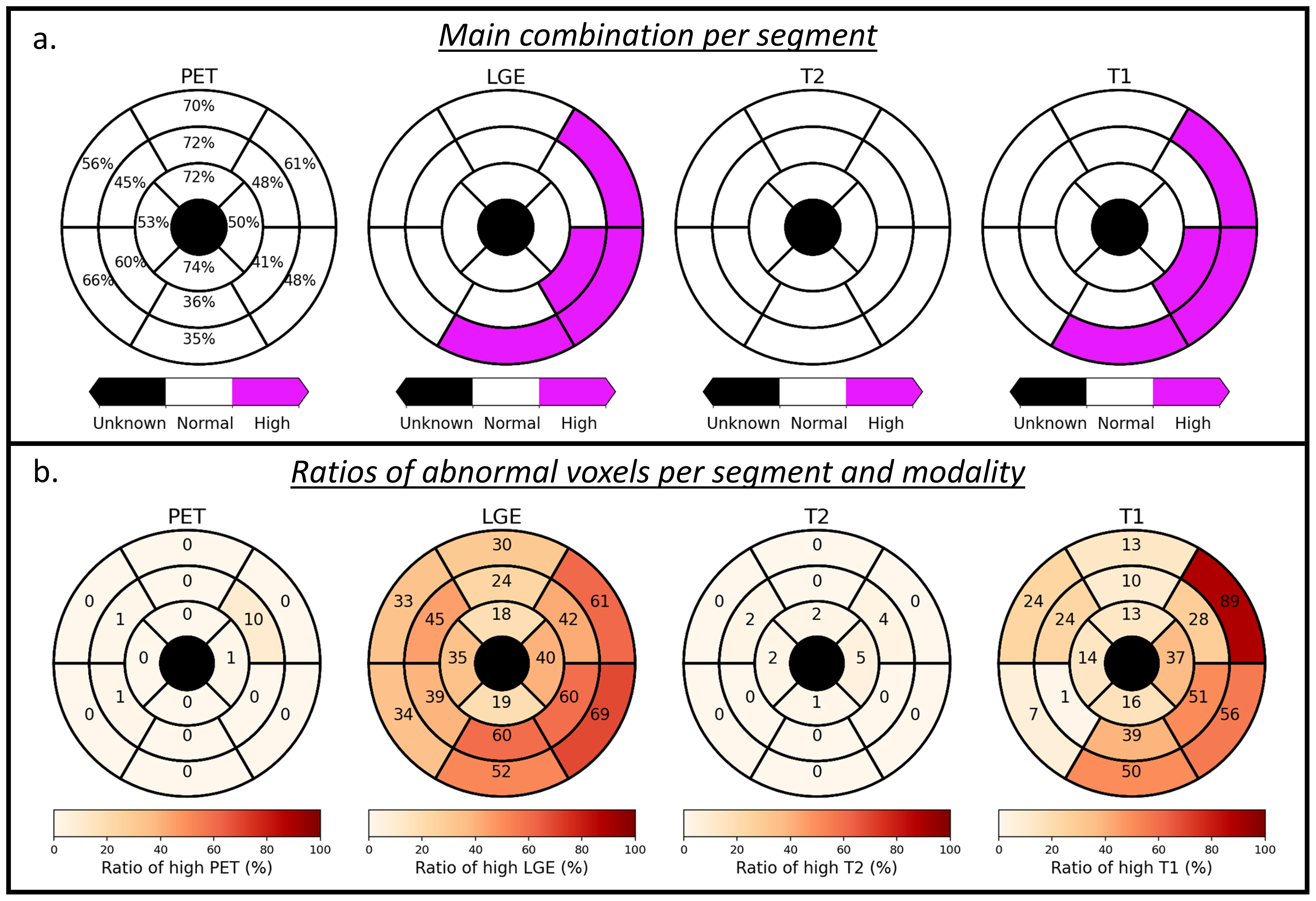}
\caption{Generated Bull's Eyes plots. In a), the main combination of the four modalities (formed by T1, T2, LGE, and PET having either normal or abnormally high voxel values) is illustrated. In b), the proportion of high voxels per segment is given per modality.}
\label{fig_5_bullseyes}
\end{figure}

To summarise the patient observations per segment in one scheme and for ease of visualisation, bull’s eye plots were created, as illustrated in Fig. \ref{fig_5_bullseyes}. This was generated using the same patient as in Fig. \ref{fig_4_results_example_visualization}. 

Two types of Bull's Eye plot were generated. For each segment, a total of  $2^4= 16$ possible combinations of T1, T2, LGE, and PET (either normal or high) could be observed. Fig. \ref{fig_5_bullseyes}a shows the main combination of the four modalities detected in the segment and its proportion. For instance, the most frequently observed combination in the anterolateral basal segment was normal PET and T2 values alongside abnormally high LGE and T1 values. This combination accounted for 61\% of the segment's total volume. Fig. \ref{fig_5_bullseyes}b illustrates the proportion of high voxels per segment in each modality using the Bull’s eye plot. Thus, the same segment was found to comprise 0\% high PET voxels, 61\% high LGE voxels, 89\% high T1 voxels, and 0\% high T2 voxels.

\section{Discussion}
This unsupervised double-step clustering method has successfully automated the process of analysing and identifying multimodal imaging combinations from PET/MRI, achieving a balanced accuracy of 0.76 in nested cross-validation on 99 patients with ALVC. It allowed for precise multimodal visualisation of abnormal zones, while summarising the reports using bull’s eye plots. This provided segment-by-segment quantification of abnormal regions across the four imaging modalities at a level of precision that would be impossible for humans to achieve in practice.

This supervoxel-based clustering workflow was inspired by several oncology studies  \cite{th_inano_visualization_heterogeneity_regional_grading_gliomas_by_multiple_features_using_MRI_clustered_images, th_tatekawa_differentiating_IDH_status_gliomas_ML_MR-PET, th_even_clustering_multiparametric_functional_imaging_to_identify_high_risk_subvolumes_lung_cancer, th_hansen_unsupervised_supervoxel-based_lung_tumor_segmentation_PET-MR} with various purposes, including glioma grading or classification \cite{th_inano_visualization_heterogeneity_regional_grading_gliomas_by_multiple_features_using_MRI_clustered_images, th_tatekawa_differentiating_IDH_status_gliomas_ML_MR-PET}, identification of tumor habitats correlated with specific outcomes \cite{th_even_clustering_multiparametric_functional_imaging_to_identify_high_risk_subvolumes_lung_cancer}, and segmentation \cite{th_hansen_unsupervised_supervoxel-based_lung_tumor_segmentation_PET-MR}. They showed the interest of using supervoxels for getting spatial information while being resistant to outliers and reducing computational costs compared to voxel-based methods. We particularly focused on the workflow proposed by \citeA{th_hansen_unsupervised_supervoxel-based_lung_tumor_segmentation_PET-MR} for segmentation, as they experimented with several variations of this two-step method. Although our final goals were different, our work can also be considered a type of segmentation task, aiming to identify separate PET/MRI habitats in the left ventricle (LV), with the additional task to identify these habitats as either abnormal or normal regions to correlate with traditional medical reports. This additional identification task was completed by defining a scoring metric attributing to each cluster abnormality score for each modality.  As T1 and T2 maps provide quantitative, medically documented values, the T1 and T2 scores were simply defined for these two modalities as piecewise linear functions centered on clinical reference values. To take advantage of the interpatient information, LGE and PET scores of each cluster were defined on the normalized PET and LGE values of the clusters by comparing the clusters to a healthy reference cluster. One challenge was thus the absence of healthy control patients, and therefore of clearly healthy clusters to use as a reference. We made the reasonable assumption that not all regions in each patient were pathological, making it possible to identify a reference cluster presumed to be healthy, against which the others could be compared. This reference cluster was built based on the hypothesis that it would not be too small, and that it would be among the clusters with the lowest PET and LGE values, while having T1 and T2 within clinically healthy ranges. Score thresholds were finally defined to dichotomize between ‘normal’ or ‘abnormal’ clusters for each modality, thus allowing for visualisation on the input volumes and comparison with the clinical routine reports. 

Additionally, some lighter modifications and additions were made to the original framework to correspond with our cardiac data and medical-related priors, while improving its robustness. Indeed, to force the clustering to focus on the LV myocardium, an extra selection of the supervoxel was added, by keeping only supervoxels with at least 20\% of their volume within the LV segmentation. This reduced the overall complexity of the model while enabling a focused identification of clusters of interest within the LV. Excluding supervoxels with less than 20\% of their surface area within the LV segmentation reduced the risk of oversegmentation by allowing the algorithm to discard poorly segmented supervoxels. The robustness of the method to imaging variations was also improved by the data augmentation techniques. The Box-Cox transformation was modified to include negative values. 

One main limitation of the proposed approach is that several methodological choices required the manual selection of parameters. Whenever possible, these parameters were validated using grid search within the nested cross-validation framework. However, some aspects, such as the exact definition of the reference cluster, could not be optimised in this way and required empirical choices. These decisions were nevertheless carefully considered and discussed in close collaboration with senior cardiologists and radiologists.

One of the challenges in this study relates to the absence of fully comprehensive ground truth data for evaluation. To address this, clustering reports were compared with segment-wise assessments provided by physicians, who manually reported the presence of high-value voxels in each modality for each cardiac segment, serving as a reference standard. While these expert annotations provide valuable clinical insight, they inherently reflect the complexity of visual interpretation, including inter-observer variability. Moreover, they primarily focus on the presence of high values within each segment, without explicitly capturing low values or intra-segment heterogeneity, thereby summarizing each segment into a single dominant state. In contrast, the proposed clustering approach offers a more detailed and quantitative characterization by estimating the distribution of multiple modality combinations within each segment. As such, this complementary perspective may provide additional granularity beyond conventional reporting, and comparisons based solely on human annotations may not fully reflect the added value of the method.

Registration between the four patient volumes was also challenging. Using supervoxels reduces the impact of small registration errors by considering only the median values of each group of voxels and excluding supervoxels with more than 80\% outside the registration area. However, poor image quality or movement meant that a few patients still had to be excluded, and large registration errors meant that in many cases, some slices or parts of slices had to be removed from the analysis. As expected, misalignments occurred most frequently in highly arrhythmogenic patients.

Finally, the limited number of patients presented a challenge in such a clustering workflow. However, using repeated nested cross-validation maximised the number of patients used while evaluating the method and its variability. Testing the method on phantoms enabled more cases to be evaluated and compared to a known gold standard. Although this model is simpler than images of real patients and is subject to fewer sources of error (such as the absence of resampling and registration steps), it enabled the method to be validated with balanced accuracy greater than 0.8. However, despite being evaluated on controlled phantom data, the model did not achieve perfect balanced accuracy. This is probably due to the Gaussian smoothing simulating partial volume effects. However, this step was necessary, as excluding the filter resulted in clustering instabilities due to numerous supervoxels with identical intensity values.

ALVC is a recently recognised pathogenic variant of ACM, which was first acknowledged with specific diagnostic criteria in the Padua criteria of 2020 \cite{Corrado_Padoue_criteria_2020}. To our knowledge, no risk score is yet defined for this specific variant, and the current diagnostic criteria still need validation on larger and more diverse cohorts \cite{th_Corrado_ACM_diagnostic_criteria_2024}. Although studies showed the interest of PET imaging \cite{th_tessier_PET_for_detection_of_myocardial_inflammation_in_ALVC, th_neves_cardiac_PET_in_ACM, th_Protonotarios_role_FDG_PET_in_ACM}, as well as T1 \cite{Everett_assessment_myocardial_fibrosis_with_T1_2016} and T2 mapping for the characterization of ALVC and fibrosis, such imaging markers are still not in the current guidelines \cite{th_Corrado_ACM_diagnostic_criteria_2024}. In this context, the proposed automatic framework enables the generation of quantitative PET/MRI reports through the identification of multimodal myocardial habitats. This approach opens perspectives for a more systematic characterization of myocardial heterogeneity in ALVC patients. It provides a basis for future work relating these habitats to distinct clinical and phenotypic patient profiles and identifying prognostic TEP/MRI biomarkers in ALVC. Combining these imaging clusters with clinical, genetic, and follow-up data could provide a better understanding of the associations between imaging and phenotype, while demonstrating the complementarity of imaging and clinical data for clinical diagnosis.

\section{Conclusion}
In conclusion, the current study focused on automating the process of generating quantitative PET/MRI imaging reports and identifying abnormal myocardial regions in ALVC patients. The proportion of high or normal PET, T1, T2, and LGE combinations for each cardiac segment of each patient was reported precisely, paving the way for future identification of connections between imaging combinations and clinical, genetic, and follow-up data. Future developments could include analysing such correlations, as well as exploring new imaging biomarkers and disease mechanisms.


\section*{Sample CRediT Author Statement}
\textbf{Brunnhilde Ponsi:} Conceptualization, Formal Analysis, Investigation,  Methodology, Software, Validation, Visualization, Writing - Original Draft. 
\textbf{Thomas Carlier:} Conceptualization, Funding Acquisition,  Methodology, Project Administration, Resources, Supervision,  Writing - Review and Editing. 
\textbf{Lara Marteau:} Methodology, Supervision, Writing - Review and Editing. 
\textbf{Aurélien Monnet:} Methodology, Writing - Review and Editing.
\textbf{Thomas Eugène:} Data Curation, Methodology, Supervision, Writing - Review and Editing. 
\textbf{Jean-Michel Serfaty:} Methodology, Supervision, Writing - Review and Editing. 
\textbf{Nicolas Piriou:} Data Curation, Methodology, Supervision, Writing - Review and Editing. 
\textbf{Hatem Necib:} Conceptualization,   Methodology, Project Administration, Resources,  Supervision,  Writing - Review and Editing. 

\section*{Declaration of Competing Interest}
Aurélien Monnet declares that he is an employee from Siemens Healthineers, which may be considered as potential competing interests with the work reported in this paper. All other authors declare that they have no known conflicts of interest in terms of competing financial interests or personal relationships that could have an influence or are relevant to the work reported in this paper.

\section*{Ethics Statement}
This work involved human subjects or animals in its research. Approval of all ethical and experimental procedures and protocols was granted by the local independent ethics committee and the clinical trial is registered under the number NCT05450783.

\section*{Data Availability Statement}
The patient data can be made available on request due to privacy/ethical restrictions.


\section*{Acknowledgments}

This work was supported by the French ”Program d’Investissement d’Avenir” (ANR-16-IDEX-0007) and the region ”Pays de la Loire” through their support to I-Site NExT, as well as by Siemens Healthineers, the industrial partner of the NExT research industrial chair IMRAM.


\appendix

\section{Choice of the SLIC parameters}\label{app1:SLIC_param}
Table \ref{Tab_parameters_slic} describes the parameters chosen for SLIC.
\begin{table}[!h]
\begin{center}
\caption{Chosen SLIC Parameters}
\label{Tab_parameters_slic}
    \begin{tabular}{ cc}
    \hline
    {Parameter} & {Value}  \\ \hline
\multicolumn{1}{l}{{n\_segments}} & $400 \cdot \text{number of slices}$ \\
\multicolumn{1}{l}{{compactness}} & 0.01 \\ 
\multicolumn{1}{l}{{max\_num\_iter}} & 10 \\ 
\multicolumn{1}{l}{{sigma}} & 0 \\ 
\multicolumn{1}{l}{{min\_size\_factor}} & 0.5 \\ 
\multicolumn{1}{l}{{max\_size\_factor}} & 3 \\ \hline
    \end{tabular}
\end{center}
\end{table}

The number of supervoxels, n\_segments, was chosen visually to delineate well-defined regions while minimizing the number of supervoxels.  The requested number of supervoxels only serves as a guideline for the algorithm, which adjusts this number as necessary to adequately segment all regions. In addition, we opted to perform SLIC within an ROI narrowed around the left ventricle, an area larger than the LV myocardium, where the supervoxels are selected and exported. Consequently, the clustering algorithm had some autonomy in determining the quantity of supervoxels to be distributed within this ROI. Compactness was intentionally set to a low value (0.01) to enable the supervoxels to adopt various shapes and align closely with potential geometric patterns of disease markers in the images. Other parameters were also explored, but the default settings of the \textit{scikit} SLIC algorithm were found to yield the best results.

\section{Data augmentation}\label{app2:data_augmentation}
Standard data augmentation techniques (rotations and flips) were used to reduce the method's sensitivity to imaging variations (Fig. \ref{fig_2_workflow}, Steps A+B). 

In addition to the initial set of images, eight modification scenarios were established, each corresponding to the application of a single alteration to the input multimodal volumes among:
\begin{itemize}
    \item rotation of 10 degrees around the y-axis
    \item rotation of 23 degrees around the x-axis
    \item rotation of 90 degrees around the x-axis
    \item rotation of 145 degrees around the z-axis
    \item rotation of 180 degrees around the y-axis
    \item flipping around the x-axis
    \item flipping around the y-axis
    \item flipping around the z-axis
\end{itemize}

Following the rotation or flipping of the input volumes, supervoxels were extracted by SLIC as described in Step B. The final set of supervoxels associated with each patient was defined as the aggregation of all supervoxels obtained across the nine scenarios.

\section{Method for validation on the phantoms}\label{app3:phantoms}
Each phantom's heart was created following the same process, but variations within the phantoms were brought by the inclusion of some randomness in the definition of their characteristics, as described in the following sections.

\subsection{Creation of a mask of the left ventricle shape}
Each phantom was given dimensions of $100 \text{ x } 100 \text{ x } Z$, where $Z$ is a randomly chosen integer between 3 and 10, representing the number of slices in which a patient can be imaged. This size was selected based on the approximate
$X \text{ x }Y$ size of the LV VOI considered when applying the intra-patient clustering on the real patients.

A LV shape was attributed to each phantom. This LV shape was modeled on each $(X,Y)$ slice as two circles inside each other, which slightly different centers (respectively $C1$ and $C2$), as human hearts are not necessarily symmetrical. 
To introduce controlled variability in the choice of the centers, a range of variation $\Delta C_1$ was defined around the image center $C_{\text{slice}_x} = C_{\text{slice}_y} = C_{\text{slice}}$. This variation was set as a fraction of the image width $X$:
\begin{equation}
\Delta C_1 = \frac{X}{20}
\end{equation}

The center \( C_1(x,y) \) of the first circle was then defined as:
\begin{equation}
\begin{cases}
C_{1x} = \text{rand} \left( C_{\text{slice}} - \Delta C_1, C_{\text{\textbf{slice}}} + \Delta C_1 \right) \\
C_{1y} = \text{rand} \left( C_{\text{slice}} - \Delta C_1, C_{\text{\textbf{slice}}} + \Delta C_1 \right)
\end{cases}
\end{equation}
where the function \( \text{rand}(a, b) \) selects a random value uniformly between \( a \) and \( b \). The second circle C2 was positioned near C1 by introducing a small random displacement around \( C_1 \). The position of \( C_2(x,y) \) was determined as follows:
\begin{equation}
\begin{cases}
C_{2x} = C_{1x} + \text{rand} \left( \frac{-\Delta C_1}{2}, \frac{\Delta C_1}{2} \right) \\
C_{2y} = C_{1y} + \text{rand} \left( \frac{-\Delta C_1}{2}, \frac{\Delta C_1}{2} \right)
\end{cases}
\end{equation}

To ensure a physiologically plausible shape variation across slices, the radii of the circles were defined to decrease progressively along the slice index. Two random scaling coefficients were defined: 
\begin{equation}
\begin{cases}
a_{\text{scaling}} = \text{rand}(0.9, 1.2)\\
b_{\text{scaling}} = \text{rand}(0.35, 0.45)
\end{cases}
\end{equation}

The coefficient \( a_{\text{scaling}} \) introduced randomness between phantoms, as a global scaling factor, while \( b_{\text{scaling}} \) was used to control the rate at which the radius decreases along the slices (as the number of slices varies).
The maximum initial radius \( R_{1,\text{max}} \) of the first circle was defined as:
\begin{equation}
 R_{1,\text{max}} = 5.2 *  \Delta C_1
\end{equation}

For each slice \( z \), the radii $R_1$ and $R_2$ of the circles \( C_1 \) and $C_2$ decreased according to the following formula:
\begin{equation}
\begin{cases}
    c_{\text{scaling}}(z) = a_{\text{scaling}} - b_{\text{scaling}} \frac{z}{(Z-1)} \\
    R_1(z) =  c_{\text{scaling}} \cdot \text{rand} \left( 4.6 \cdot\Delta C_1,R_{1,\text{max}} \right) \\
    R_2(z) = R_1(z) - \text{rand} \left( \frac{4 \cdot\Delta C_1}{3}, 2\cdot\Delta C_1 \right)
\end{cases}
\end{equation}
where \( Z \) is the total number of slices. 

A mask of the LV was finally defined as the voxels in between the two circles.

\subsection{Random creation of unhealthy LGE, T1, T2 and PET zones}
Five distinct masks were defined to represent the localization of randomly generated abnormal regions in LGE, T1, T2, PET, and 'PET - large' imaging. The large PET mask was introduced to account for two different types of abnormal PET signal variations, as PET abnormalities tend to exhibit greater variability in human subjects compared to MRI-based markers.

For each phantom, a random number of abnormal zones were created, uniformly chosen between 0 and 30. Each region was constructed following the procedure below: 
\begin{enumerate}
    \item A random point $(x,y,z)$ within the LV is selected.
    \item A circular mask centered at the selected point 
$(x,y,z)$ is defined on the corresponding slice $z$, representing the locality of a lesion in the myocardium. The radius of this mask is randomly selected from the interval $[\Delta C_1, \frac{5\cdot\Delta C_1}{2}]$. For large PET regions, a larger radius is applied in $ 2.5\cdot[\Delta C_1, \frac{5\cdot\Delta C_1}{2}]$, reflecting the typically greater extent of PET abnormalities compared to MRI-detected lesions. 
    \item A second ring-shaped mask is then created to simulate a radial distribution of abnormalities around the LV. It is composed of a $(X,Y)$-plane ring of center $C_1$, width comprised between 0 and 5 voxels, and passing through $(x,y,z)$.
    \item The product of these two masks is computed to retain only the overlapping region. This ensures that the anomaly is both spatially constrained around the selected point and aligned with a concentric pattern around the LV, mimicking physiologically plausible distributions of myocardial lesions, such as fibrosis or infarction.
    \item Finally, this constructed abnormal zone is attributed to the five abnormal LGE, T1, T2, PET,  and large PET masks with various probabilities:
    \begin{enumerate}
        \item 70\% probability of inclusion in the LGE abnormality mask.
        \item 40\% probability of inclusion in the T1 abnormality mask.
        \item 20\% probability of inclusion in the T2 abnormality mask.
        \item 40\% probability of inclusion in the PET abnormality mask.
        \item 50\% probability of inclusion in the 'PET - large' abnormality mask.
    \end{enumerate}
\end{enumerate}

\subsection{Creation of the MRI and PET volumes, and inclusion of noise}
T1, T2, LGE, and PET volumes were then created by differentiating three regions using the mask of the LV previously created: the ventricular cavity, the myocardium, and the area outside the LV.  The inclusion of various levels of noise was tested.

Each of these regions was given voxels values based on a "central" value $X_{mod}^{region}$. The expression of these voxel values can be divided into two components:
\begin{enumerate}
    \item First, a uniform value was given to the region, equal to a random number within $[(1  - \eta) \cdot X_{mod}^{region}, (1+ \eta) \cdot X_{mod}^{region}]$, where $\eta$ is the noise percentage. This value allowed variation between phantoms.
    \item Secondly, random uniform noise in $[-\eta X_{mod}^{region}, \eta X_{mod}^{region}]$ was added to each voxel of the zone to simulate variability within voxels.
\end{enumerate}

The values chosen for the central values per region are given in Table \ref{tab_central_values_creation_phantoms}.

\begin{table}[!h]
\begin{center}
\caption{Central Values $X_{mod}^{region}$ Attributed to Each Region for Each Modality in the Creation of the Phantom Volumes.}
\label{tab_central_values_creation_phantoms}
    \begin{tabular}{ ccll}
    \hline
    {Sequence}& {Ventricular Cavity}& {Myocardium}&{Outside the LV}\\ \hline
\multicolumn{1}{l}{{PET}} & 2500& 3000&2000\\ 
\multicolumn{1}{l}{{LGE}} & 2222& 2075&2222\\ 
\multicolumn{1}{l}{{T1}} & 1800& 1250&1000\\
\multicolumn{1}{l}{{T2}} & 70& 43&130\\ 
    \end{tabular}
\end{center}
\end{table}

Finally, the abnormal regions identified in the MRI and PET masks were incorporated into the modality volumes by modifying the previously defined voxel values. For each voxel within an abnormal area, a random value was added, sampled from the range $[(1  - \eta) \cdot Y_{mod}, (1+ \eta) \cdot Y_{mod}]$, where $Y_{mod}$ represents a predefined abnormality-related additive term specific to each modality, and is given in Table \ref{tab_addition_abnormal_values_creation_phantoms}.

\begin{table}[!h]
\caption{Abnormality-Related Additive Terms $Y_{mod}$ used in the Creation of the Phantom Volumes.}
\label{tab_addition_abnormal_values_creation_phantoms}
\begin{center}
    \begin{tabular}{ cc}
    \hline
    {Sequence}& {$Y_{mod}$}\\ \hline
\multicolumn{1}{l}{{PET}} & 750\\ 
\multicolumn{1}{l}{{LGE}} & 200\\ 
\multicolumn{1}{l}{{T1}} & 100\\
\multicolumn{1}{l}{{T2}} & 6\\ 
    \end{tabular}
\end{center}
\end{table}

As two different masks represented abnormal areas in PET, abnormal values from both masks were successively added to the PET volume of each phantom.

\subsection{Inclusion of the volume partial effect }
A 3D Gaussian filter was finally applied to each of the created volumes to simulate the volume partial effect. The filter was applied in the $x$, $y$, and $z$ directions by adjusting the kernel size to account for anisotropic voxel spacing. The width of the filter was chosen to be 2 mm for MRI volumes and 4 mm for PET volumes.

\subsection{Example of one numerical phantom}
\begin{figure}[!t]
\centering
\includegraphics[width=1\linewidth]{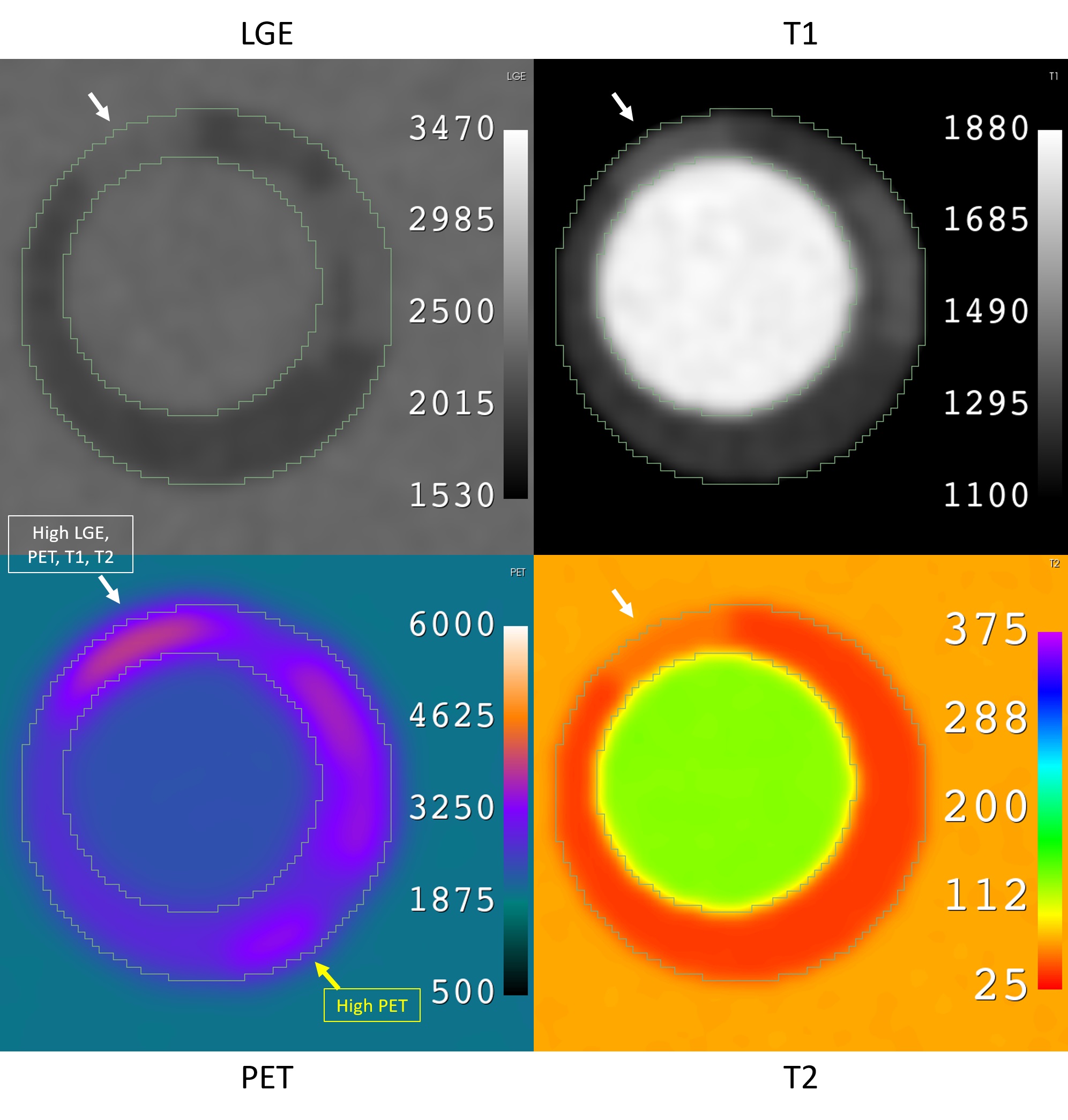}
\caption{Example of one randomly generated phantom with 5\% noise.}
\label{fig_example_one_phantom}
\end{figure}

Fig. \ref{fig_example_one_phantom} gives an example of one randomly generated phantom with 5\% noise.

\section{Choice of hyperparameters}\label{app4:hyperparameters}
The nested cross‑validation relied on an inner grid search with conditional hyperparameter sets, defined as follows.
When affinity was "nearest\_neighbors":
\begin{itemize}
    \item $\text{n\_neighbors} \in \{5, 10, 12, 15, 20, 30\}$
    \item $\text{thresh\_high\_LGE} \in \{0.15, 0.20, 0.25\}$
    \item $\text{thresh\_high\_PET} \in \{0.15, 0.20, 0.25\}$
\end{itemize}
When affinity was  "polynomial":
\begin{itemize}
    \item $\text{thresh\_high\_LGE} \in \{0.15, 0.20, 0.25\}$
    \item $\text{thresh\_high\_PET} \in \{0.15, 0.20, 0.25\}$
\end{itemize}
When affinity was "rbf" or "laplacian":
\begin{itemize}
    \item $\text{gamma} \in \{0.15, 0.25, 0.35, 0.50, 0.60, 0.75, 1, 5, 10\}$
    \item $\text{thresh\_high\_LGE} \in \{0.15, 0.20, 0.25\}$
    \item $\text{thresh\_high\_PET} \in \{0.15, 0.20, 0.25\}$
\end{itemize}
with thresh\_high\_LGE and thresh\_high\_PET the thresholds used to distinguish between normal and abnormally high score values in a cluster.

For visualization purposes and tests on the phantoms, hyperparameters had to be selected. The Laplacian affinity was selected, as it was chosen in 90\% of cases during cross-validation. When this kernel was selected, the thresholds for high PET and high MRI scores were both equal to 0.25 in more than 50\% of cases (79\% and 53\%, respectively), and were therefore retained as the optimal hyperparameters. In contrast, the gamma parameter, corresponding to the kernel coefficient, showed greater variability, most frequently ranging between 0.15 and 0.35, as shown in Table \ref{tab_distrib_gamma_nCV}. It was thus fixed to 0.25. When this kernel was fixed, n\_neighbors was not a parameter.

\begin{table}[!h]
\begin{center}
\caption{Distribution of the Percentage of Time Each Gamma Value was Identified as the Best Hyperparameter in the Nested Cross-Validation, When Considering Cases with Laplacian Kernel. 
}
\label{tab_distrib_gamma_nCV}
    \begin{tabular}{ cc}
    \hline
    {Gamma} & {Percentage of time}  \\ \hline
\multicolumn{1}{l}{{0.15}} & 33\% \\
\multicolumn{1}{l}{{0.25}} & 20\% \\ 
\multicolumn{1}{l}{0.35} & 26\% \\ 
\multicolumn{1}{l}{{0.5}} & 11\% \\ 
\multicolumn{1}{l}{0.6} & 5\% \\ 
\multicolumn{1}{l}{0.75} & 3\% \\ 
\multicolumn{1}{l}{1} & 3\% \\ \hline
    \end{tabular}
\end{center}
\end{table}

\bigbreak
\FloatBarrier
\bibliographystyle{elsarticle-num-names-6}
\bibliography{mylib}





\end{document}